\documentclass[10pt,twocolumn,letterpaper]{article}

\usepackage{wacv}
\usepackage{times}
\usepackage{epsfig}
\usepackage{graphicx}
\usepackage{amsmath}
\usepackage{amssymb}
\usepackage{multirow}
\usepackage{multicol}
\usepackage{booktabs}
\usepackage{authblk}
\usepackage{blindtext}
\usepackage[skip=2pt]{caption}
\usepackage[symbol]{footmisc}
\def\wacvPaperID{35} % Enter the WACV Paper ID here

\wacvfinalcopy % *** Uncomment this line for the final submission

\ifwacvfinal
\usepackage[breaklinks=true,bookmarks=false]{hyperref}
\else
\usepackage[pagebackref=true,breaklinks=true,colorlinks,bookmarks=false]{hyperref}
\fi

\ifwacvfinal
\fi
\begin{document}

%%%%%%%%% TITLE
\title{Scale Aware Adaptation for Land-Cover Classification\\ in Remote Sensing Imagery}

\author[1]{Xueqing Deng}
\author[2]{Yi Zhu\thanks{Work was done prior to joining Amazon.}} 
\author[1]{Yuxin Tian}
\author[1]{Shawn Newsam}
\affil[1]{EECS, UC Merced \textit {\{xdeng7, ytian8, snewsam\}@ucmerced.edu}}
\affil[2]{Amazon Web Services \textit {\{yzaws\}@amazon.com}}
% \affil[1]{\textit {\{xdeng7,ytian8,snewsam\}@ucmerced.edu}}
% \affil[2]{\textit {\{yzaws\}@amazon.com}}
% \footnote[1]{Work was done prior to Amazon.}
\maketitle
%\thispagestyle{empty}

% \addtolength{\parskip}{-0.3mm}

%%%%%%%% ABSTRACT
\begin{abstract}
\vspace{-10pt}
Land-cover classification using remote sensing imagery is an important Earth observation task. Recently, land cover classification has benefited from the development of fully connected neural networks for semantic segmentation. The benchmark datasets available for training deep segmentation models in remote sensing imagery tend to be small, however, often consisting of only a handful of images from a single location with a single scale. This limits the models' ability to generalize to other datasets. Domain adaptation has been proposed to improve the models' generalization but we find these approaches are not effective for dealing with the scale variation commonly found between remote sensing image collections. We therefore propose a scale aware adversarial learning framework to perform joint cross-location and cross-scale land-cover classification. The framework has a dual discriminator architecture with a standard feature discriminator as well as a novel scale discriminator. We also introduce a scale attention module which produces scale-enhanced features. Experimental results show that the proposed framework outperforms state-of-the-art domain adaptation methods by a large margin. The open-sourced codes are available on Github: \url{ https://github.com/xdeng7/scale-aware_da}. 
\vspace{-15pt}
\end{abstract}

\section{Introduction}
\label{sec:introduction}
% \vspace{-5pt}

High-resolution remote sensing imagery is becoming increasingly available due to the number and revisit rate of traditional satellite and airborne image capture platforms as well as the advent of newer platforms like drones. This imagery provides convenient and large-scale coverage and so is being applied to a number of societally important problems such as land cover segmentation \cite{Kussul_2017_LULC}, traffic monitoring \cite{Ma_2017_traffic}, urban planning \cite{Ball_2017_survey}, vehicle detection \cite{Chen_2014_car}, building extraction \cite{Yuan_2017_building}, geolocolization\cite{Tian_2020_WACV} etc. While remote sensing (RS) image analysis has benefited from advances in deep learning in the computer vision community, there often remains unique challenges that limit the straightforward application of standard approaches to the RS case.

\begin{figure}[t]
    \centering
    \includegraphics[width=0.7\linewidth]{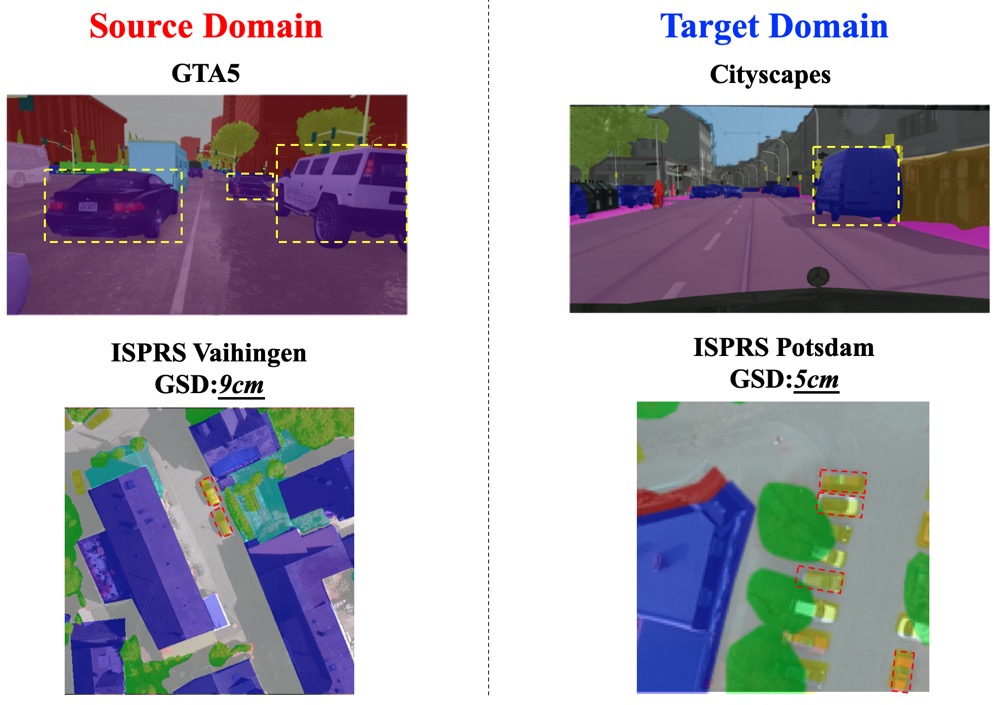}
    % \vspace{0.5ex}
    \setlength{\belowcaptionskip}{-18pt}
    \caption{Object sizes vary both within and between ground-level image datasets such as in the driving scenes at the top. Segmentation models trained on one dataset are already scale-invariant and so standard domain adaptation techniques are not designed to explicitly account for scale variation. In contrast, object sizes usually do not vary within RS image datasets since all the images have the same ground sample distance (GSD). We propose a novel framework that explicitly accounts for scale when adapting between RS image datasets with different scales such as at the bottom.}

    \label{fig:overview}
\end{figure}

Another issue that limits the performance of automated RS image analysis, particularly deep learning approaches, is that the availability of the annotated ground truth data needed for training has not kept pace with the imagery (or its diversity). As an example, current semantic segmentation datasets for land cover classification, which are very labor intensive to produce, contain limited labeled samples from only a few locations. The ISPRS Vaihingen dataset \cite{isprs_challenge} contains just $33$ labeled tiles with 6 semantic classes. The recent SkyScapes \cite{azimi_iccv_skyscapes} dataset has more classes with 30 but still contains only $16$ images. DeepGlobe \cite{demir2018deepglobe} is the largest collection, containing $1146$ images with 7 classes. Deep learning models trained on these (relatively) small datasets have difficulty generalizing to other image sets, i.e. large-scale WorldView imagery. They suffer from the so-called domain shift problem \cite{chen2017iccv,tsai2018learning}. One of the great benefits of using RS imagery is its fast revisit time and large coverage for Earth observation. However, this scalability is limited by domain shift problems.

Domain shift in RS imagery can happen along a number of dimensions including differences in sensor, spectra, resolution, etc. which have been widely explored \cite{remotessensing2019,spetralVolpi2018,bosch2019sensor}. There has been much less work, somewhat surprisingly, on the cross-location domain shift problem in which a model trained on one location that has ground truth data, the source location, is applied to another location without ground truth data, the target location. The work that has been done simply applies standard domain adaptation techniques \cite{remotessensing2019,siameseicip2018,uda2019deng}. However, none of this work explicitly considers the important notion of scale which we believe is often integral to the cross-location problem especially for tasks where there is limited training data like semantic segmentation in RS imagery. We therefore propose a novel scale adaptive framework for improved cross-location domain adaptation. The framework is adversarial and includes separate feature and scale discriminators as well as a scale attention module.

Domain shift caused by location also exists in other datasets such as road scene imagery. However, we found that scale is more important for overhead (RS) than (most) ground-level imagery. A simple visual investigation illustrates why scale is more important. Fig.\ \ref{fig:overview} contains a pair of images from different driving scene datasets and a pair of images from different RS image datasets. (The ground truth segmentation masks are overlaid on the images.) The objects in the road scenes vary in size both within a dataset (even within an image) as well as between datasets. A deep learning model trained on one dataset learns scale-invariant features so is better able to generalize to the other dataset. 
% (if it is meaningful to even consider what scale means here). 
However, in the RS imagery, due to the planar nature of the scene and the nadir viewpoint, the size of the objects does not vary (much) within a dataset if all the images have the same resolution or ground sample distance (the physical size of a pixel on the ground typically indicated in meters or similar) which is usually the case. The RS imagery in the source domain in Fig.\ \ref{fig:overview} has a GSD of 9cm so all the cars measure around $17\times22$ pixels assuming $1.5\times2$ meter cars. The GSD of the target domain is just 5cm so all the cars are proportionally larger and measure around $30\times40$ pixels. A model trained on one dataset will not learn the scale-invariant features needed to label the differently sized cars in the other dataset. Data augmentation during training through random scaling is not an effective solution. Additional examples of showing that scale is more important for RS than regular imagery can be found in the supplementary material.
% To demonstrate this quantitatively, we compare the robustness of standard domain adaptation techniques to changes in scale in overhead and ground-level imagery. We train a road scene segmentation model on the GTA5 dataset and then adapt this model to the Cityscapes dataset \cite{Cordts2016Cityscapes}. We then apply the model to scaled versions of the Cityscapes test images. The yellow bars in Figure \ref{fig:aerial_road} show the drop in performance for downsampling by factors of 2 and 4. We similarly train a land cover segmentation model on the ISPRS Vaihingen dataset \cite{isprs_challenge}, adapt the model to the ISPRS Potsdam dataset \cite{isprs_challenge}, and apply the model to downsampled versions of the ISPRS Potsdam test images. The blue bars show the corresponding drop in performance. Changing the scale of the overhead imagery results in a bigger drop in performance. In the 1/4 scaled case, the land cover segmentation drops by more than $20\%$, while the road scene segmentation is only slightly worse. This experiment demonstrates that scale is an important factor to consider in overhead image domain adaptation.

We therefore develop a novel domain adaptation framework that explicitly accounts for scale changes in order to improve generalization in cross-location semantic segmentation for RS imagery. 
The proposed framework contains dual adversarial discriminators including a standard feature discriminator and a novel scale discriminator as well as a scale attention module. There are separate discriminators for feature and scale adaptation between the source and target domains. The scale attention module selectively weights concatenated multi-scale features to obtain scale-enhanced features.
Our novel contributions include:
\vspace{-4pt}
\begin{itemize}
    \item We establish and demonstrate that explicitly accounting for scale variation is integral to RS image domain adaptation yet no existing work does this. We propose an innovative scale adaptive framework for cross-location semantic segmentation which includes a novel scale discriminator and a scale attention module to facilitate training. We demonstrate that scale-aware adaptation results in significant performance gains.
    \vspace{-5pt}
    \item We show our proposed approach outperforms state-of-the-art domain adaptation methods on several cross-dataset segmentation tasks such as Potsdam $\leftrightarrow$ Vaihingen and DeepGlobe $\leftrightarrow$ Vaihingen.
\end{itemize}
% \vspace{-10pt}
\section{Related Work}
\label{sec:relatedwork}
% \vspace{-5pt}
\noindent \textbf{Semantic segmentation}
There is a large body of literature on semantic segmentation\cite{chen2017deeplabv2, chen2018deeplabv3plus, fu2019danet, zhu2019cvpr}. Here, we review only the most related work in terms of handling scale \cite{zhao2017pspnet,ding2018context,lin2018multi,jin2019casenet}. Fully Convolutional Network (FCN) based methods \cite{long2015fully} have made great progress in semantic segmentation. To capture multi-scale information, DeepLab networks \cite{chen2017deeplabv2}, PSPNet \cite{zhao2017pspnet} and CaseNet \cite{jin2019casenet} develop mechanisms to integrate multi-scale features. Benefiting from the above exploration on regular images, a number of works \cite{casnet2018isprs,Mnih_2010_eccv_road,nicolas2018isprs,dlinknet2018cvpr,deng2019generalizing, deng2018like} have applied the techniques to pixel-wise land-cover classification in RS imagery. These methods focus on multi-scale analysis but not explicitly on scale adaptation as proposed in this paper.

\noindent \textbf{Domain adaptation}
Numerous domain adaptation methods have been developed to improve image classification by aligning the feature distributions between the source and the target domains \cite{tzeng2017adversarial,ganin2016dann,murez2018image,aannTGARSL2018}. Domain adaptation for segmentation has recently started to receive more attention due to the expense of performing pixel-wise annotation. Many approaches have been proposed for road scene segmentation \cite{tsai2018learning,chen2018road,hoffman2018cycada,hoffman2016fcns,tsai2019domain,vu2019dada,kim2020learning,zou2019CRST,chen2019cvpr}. Since scale is much less significant for road scene images, these studies focus mainly on adapting texture, appearance, etc. There has been some work on using Generative Adversarial Networks (GAN) for domain adaptation in RS image segmentation \cite{siamenseGAN2019rs,siameseicip2018}. However, these approaches just apply standard computer vision methods without considering challenges specific to RS imagery. We instead propose a framework that addresses the importance of scale when adapting between domains.

\noindent \textbf{Attention}
Attention was originally proposed to exploit long-range dependencies in machine translation \cite{vaswani2017attention}. It has since been adopted in a number of computer vision tasks \cite{zhang2018rcan,wang2018nlnn,zhang2019sagan,zhu2019ann}. Self-attention has been used as a non-local operation to learn positional relations in video classification \cite{wang2018nlnn}, to learn a better image generator \cite{zhang2019sagan}, as well as to learn both channel and spatial attention for scene segmentation \cite{fu2019dual} and land cover segmentation \cite{mou2019relation}. In order to augment the extraction of multi-scale information, we propose a scale attention module (channel attention) which improves the scale adaptation.

% \vspace{-5pt}
\section{Problem Formulation}
\label{sec:problem}
% \vspace{-5pt}
We formulate our problem as cross-scale and cross-location domain adaptation for semantic segmentation in RS imagery (pixel-wise land-cover classification). We assume the domain shift is caused by different scales and locations between the source and target datasets. We recognize, though, that different locations do not necessarily have different scales. Our framework is unified in that it can deal with domain shift caused by scale or by location or by both.

We denote the different locations as $\mathcal{S}$ and $\mathcal{T}$ and the different scales as $\theta$ and $\sigma$. We denote an image ${x}$ from source location $\mathcal{S}$ with scale $\theta$ as ${x}_{\mathcal{S}^{\theta}} \in {X\{location=\mathcal{S},scale=\theta\}}$ and its label as ${y}_{\mathcal{S}^{\theta}} \in $ ${{{Y\{location=\mathcal{S},scale=\theta\}}}}$. $X$ and $Y$ represent all the images and labels in one domain. Our goal is to adapt, in an unsupervised manner, a segmentation model $\mathcal{G}$ trained using images ${X}_{\mathcal{S}^{\theta}}$ and labels ${Y}_{\mathcal{S}^{\theta}}$ from source domain $\mathcal{S}^{\theta}$ to perform segmentation in target domain $\mathcal{T}^{\sigma}$ which has only images ${X}_{\mathcal{T}^{\sigma}}$ to produce predicted segmentation masks $\hat{{Y}_{\mathcal{T}^{\sigma}}}$.

% \vspace{-5pt}
\section{Methods}
\label{sec:method}
% \vspace{-5pt}
In this section, we describe our scale aware domain adaptation framework. We first revisit conventional domain adaptation methods which only have feature discriminators. We then describe our new scale discriminator for addressing the scale variation between domains. Finally, we describe our scale attention module for obtaining scale-enhanced features for improved segmentation.
\begin{figure*}[t]
\setlength{\abovecaptionskip}{-2pt}
    \setlength{\belowcaptionskip}{-12pt}
    \centering
    \includegraphics[width=0.9\textwidth]{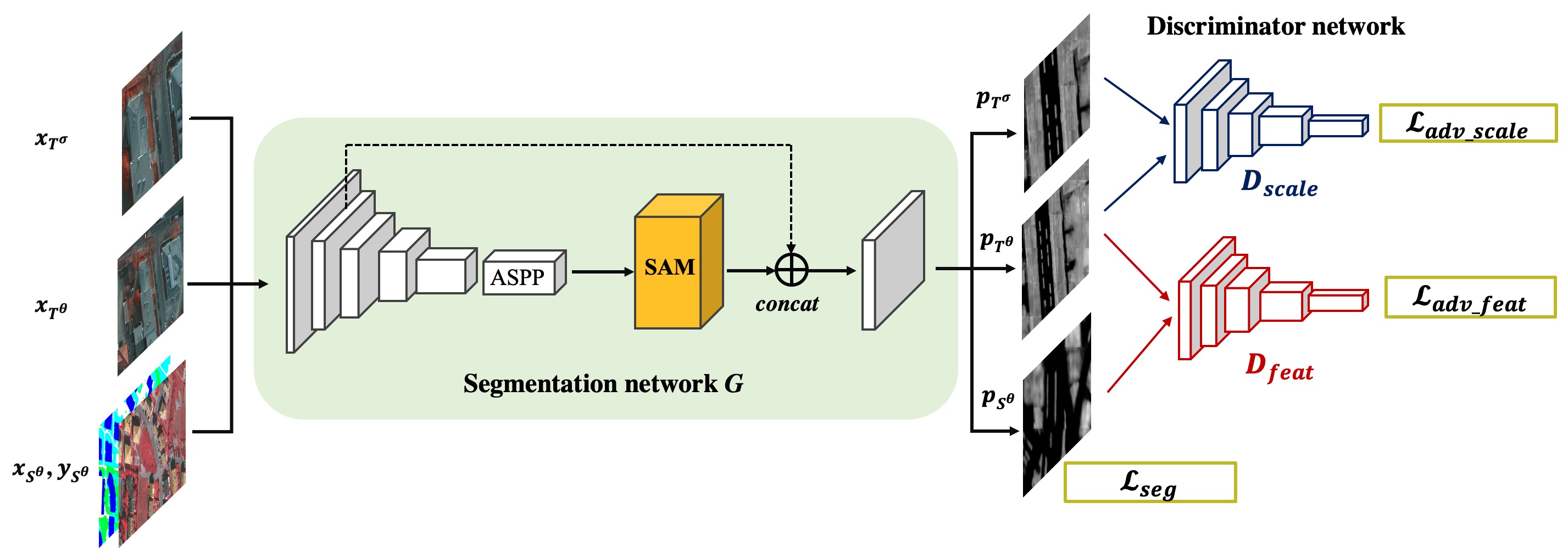}
    \caption{Our proposed scale aware adaptation framework which contains two adversarial learning components, one for feature adaptation and another for scale adaptation. We use DeepLabV3+ as our segmentation network. ASPP is an Atrous Spatial Pyramid Pooling module. We also propose a scale attention module (SAM) to produce scale-enhanced features. See the text for more details.}

\label{fig:framework}
\end{figure*}

% \vspace{-5pt}
\subsection{Domain Adaptation Revisited}
\vspace{-5pt}
\label{subsec:revisit}
Recent domain adaptation methods for semantic segmentation are adversarially based \cite{tsai2018learning,hoffman2018cycada}. An image from either the source or target domain is input to the segmentation network. The resulting feature maps or segmentation predictions are then fed to a discriminator which tries to determine the domain of the input. The goal of the segmentation network now is to not only produce an accurate segmentation of the source domain image (for which we have labels) but to also fool the discriminator. This forces the internal representations of the model to align between domains so it can better exploit its supervised training in the source domain when it is applied to the target domain.

The segmentation network is now updated using two losses, the segmentation loss when the input is from the source domain and the adversarial loss when the input is from the target domain. Given any segmentation network $\mathcal{G}$ (e.g., \cite{zhao2017pspnet,chen2018deeplabv3plus}), source image ${x}_{\mathcal{S}^{\theta}}$ will result in prediction ${p}_{\mathcal{S}^{\theta}} = \mathcal{G} ({x}_{\mathcal{S}^{\theta}})$ and target image ${x}_{\mathcal{T}^{\sigma}}$ will result in prediction ${p}_{\mathcal{T}^{\sigma}} = \mathcal{G} ({x}_{\mathcal{T}^{\sigma}})$. Note that we choose to use the low-dimensional softmax output predictions instead of the feature maps following \cite{tsai2018learning} since they contain rich spatial information shared between locations which makes it easier for the network to adapt. When the input is from the source domain, the multi-class cross-entropy segmentation loss 
\begin{equation}
\setlength{\abovedisplayskip}{5pt}
\setlength{\belowdisplayskip}{5pt}
    \mathcal{L}_{seg} ({x}_{\mathcal{S}^{\theta}}) = -  {y}_{\mathcal{S}^{\theta}}
    \log ({p}_{\mathcal{S}^{\theta}}),
    \label{eq:segloss}
\end{equation}
\noindent is computed where ${y}_{\mathcal{S}^{\theta}}$ is the annotated label.
A segmentation loss cannot be computed for a target image since its label is not available. So, in order to adapt the segmentation network to the target domain, a feature discriminator $\mathcal{D}_{feat}$ is added and an adversarial loss is calculated
\begin{equation}
\setlength{\abovedisplayskip}{5pt}
\setlength{\belowdisplayskip}{5pt}
    \mathcal{L}_{adv\_feat}({x}_{\mathcal{T}^{\sigma}}) =  -\log{(\mathcal{D}_{feat}({p}_{\mathcal{T}^{\sigma}})}).
    \label{eq:adv_feat_loss1}
\end{equation}
\noindent This is a binary cross-entropy loss, designed to fool the discriminator by forcing the space of target predictions ${p}_{\mathcal{T}^{\sigma}}$ to match the space of source predictions ${p}_{\mathcal{S}^{\theta}}$. At this point, the overall objective for updating the segmentation network is a combination of $\mathcal{L}_{seg}$ and  $\mathcal{L}_{adv\_feat}$
\begin{equation}
\setlength{\abovedisplayskip}{5pt}
\setlength{\belowdisplayskip}{5pt}
    \mathcal{L}({x}_{\mathcal{S}^{\theta}},{x}_{\mathcal{T}^{\sigma}}) = \mathcal{L}_{seg} + \lambda_{f}\mathcal{L}_{adv\_feat},
    \label{eq:sum_loss1}
\end{equation}
\noindent where $\lambda_{f}$ is the weight of the adversarial loss. 

The feature discriminator $\mathcal{D}_{feat}$ is updated using its own adversarial loss 
\begin{equation}
\setlength{\abovedisplayskip}{5pt}
\setlength{\belowdisplayskip}{5pt}
\begin{split}
    \mathcal{L}_{\mathcal{D}_{feat}}({p}) = - {(1-z)}  \log(\mathcal{D}_{feat}(p)) +z\log(\mathcal{D}_{feat}(p))
\end{split}
\label{eq:dfeat_loss1}
\end{equation}
where $z=0$ if the output prediction map $p$ is from the target location $\mathcal{T}$, and $z=1$ if it is from the source location $\mathcal{S}$. The segmentation network and the feature discriminator are optimized in an alternating manner. When one is being updated, the other is frozen.
% \vspace{-5pt}
\subsection{Scale Discriminator}
\vspace{-5pt}
\label{subsec:sdiscriminator}

The standard domain adaption framework above achieves decent results for cross-location segmentation when the 
source and target domain have similar scale, i.e.,  ${x}_{\mathcal{S}^{\theta}}$ and ${x}_{\mathcal{T}^{\theta}}$. However, it does not do well when the scale varies, i.e., ${x}_{\mathcal{S}^{\theta}}$ and ${x}_{\mathcal{T}^{\sigma}}$. When the scale of target dataset is different from the source, the performance of a model trained with just a feature discriminator decreases by $20\%$ in the RS case (details in the supplementary material).

We therefore propose a dual discriminator network which includes the standard feature discriminator as well as a new scale discriminator. We split the adaptation task into two sub-tasks, one that focuses on (cross-location) feature adaptation and another that focuses on scale adaptation. The scale discriminator has the same network structure as the feature discriminator. The framework is shown in Fig.~\ \ref{fig:framework}.

We now have three kinds of input images instead of two. Besides the source ${x}_{\mathcal{S}^{\theta}}$ and target ${x}_{\mathcal{T}^{\sigma}}$ images, we derive a resized version of the target image ${x}_{\mathcal{T}^{\theta}}$ via bilinear interpolation whose scale matches that of the source. These three types of inputs allow us to create two adversarial flows, one that considers images from the same location but different scales, and another that considers images with the same scale but from different locations. The key to our framework is the new, matched scale image ${x}_{\mathcal{T}^{\theta}}$. It allows our network to focus on adapting features between the source and target locations without also having to adapt for scale.

The feature discriminator $\mathcal{D}_{feat}$ is now updated using images with the same scale but from different locations using the adversarial loss 
\begin{equation}
\setlength{\abovedisplayskip}{5pt}
\setlength{\belowdisplayskip}{5pt}
\begin{split}
    \mathcal{L}_{\mathcal{D}_{feat}}({p_{\theta}}) = - {(1-z)}  \log(\mathcal{D}_{feat}(p_{\theta})) +z\log(\mathcal{D}_{feat}(p_{\theta})).
\end{split}
\label{eq:dfeat_loss2}
\end{equation}
Note that the difference between (\ref{eq:dfeat_loss1}) and (\ref{eq:dfeat_loss2}) is that the output prediction $p$ in (\ref{eq:dfeat_loss2}) has the same scale $\theta$ no matter whether it is from the source or target location.

The scale discriminator $\mathcal{D}_{scale}$ is updated using images from the same location but with different scales using the adversarial loss
\begin{equation}
\setlength{\abovedisplayskip}{5pt}
\setlength{\belowdisplayskip}{5pt}
\begin{split}
    \mathcal{L}_{\mathcal{D}_{scale}}({p}_{\mathcal{T}}) = - {(1-z)}\log(\mathcal{D}_{scale}({p}_{\mathcal{T}}) +z\log(\mathcal{D}_{scale}({p}_{\mathcal{T}})),
\end{split}
\label{eq:sfeat_loss1}
\end{equation}
where $z=0$ if $p_{\mathcal{T}}$ is has the target scale $\sigma$, and $z=1$ if it has the source scale $\theta$. 

We now turn to the update of the segmentation network. Similar to (\ref{eq:adv_feat_loss1}), a feature adversarial loss is calculated using the feature discriminator to adapt the segmentation network to the target features 
\begin{equation}
\setlength{\abovedisplayskip}{5pt}
\setlength{\belowdisplayskip}{5pt}
    \mathcal{L}_{adv\_feat}({x}_{\mathcal{T}^{\theta}}) =  -\log{(\mathcal{D}_{feat}({p}_{\mathcal{T}^{\theta}})}).
    \label{eq:adv_feat_loss2}
\end{equation}
In order to adapt the segmentation network to the target scale, a scale adversarial loss is computed as
\begin{equation}
\setlength{\abovedisplayskip}{5pt}
\setlength{\belowdisplayskip}{5pt}
\begin{aligned}
        \mathcal{L}_{adv\_scale}({x}_{\mathcal{T}^{\sigma}})= -\log (\mathcal{D}_{scale}( {p}_{\mathcal{T}^{\sigma}}).
\end{aligned}
\label{eq:adv_scale_loss1}
\end{equation}
The overall objective for updating the segmentation network is the sum of the three losses,
\begin{equation}
\setlength{\abovedisplayskip}{5pt}
\setlength{\belowdisplayskip}{5pt}
    \mathcal{L}({x}_{\mathcal{S}^{\theta}}, {x}_{\mathcal{T}^{\sigma}}) = \mathcal{L}_{seg} + \lambda_{f}\mathcal{L}_{adv\_feat} +\lambda_{s}\mathcal{L}_{adv\_scale}.
    \label{eq:sum_loss2}
\end{equation}
Here, $\lambda_s$ and $\lambda_f$ are the hyperparameters for loss weights. 

\begin{figure}[t]
\setlength{\abovecaptionskip}{-2pt}
    \setlength{\belowcaptionskip}{-8pt}
    \centering
    \includegraphics[width=\linewidth]{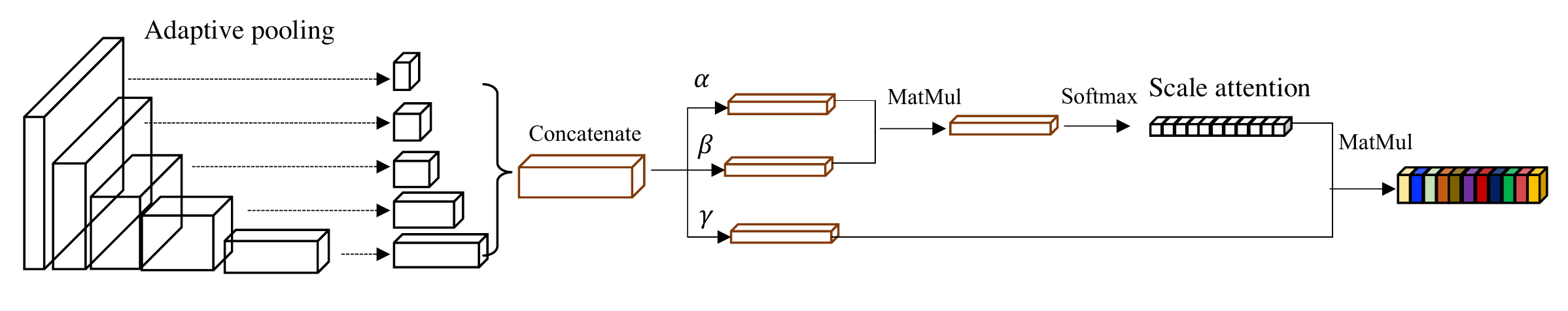}
    \caption{Proposed scale attention module (SAM). The input is feature maps from the segmentation network. The output is scale-enhanced features. MatMul: matrix multiplication.}
     \label{fig:attention}
       \vspace{-5pt}

%    \caption{Scale Attention Module (SAM). The input contains features extracted from different layers of a segmentation network. A channel self-attention is then performed on the multi-scale concatenated feature. MatMul: matrix multiplication.}   
\end{figure}

\begin{table*}[t!]
\setlength{\abovecaptionskip}{-2pt}
    \setlength{\belowcaptionskip}{-3pt}
  \centering
  \caption{Comparing our approach which incorporates scale adaptation to resampling the target imagery with and without standard domain adaptation. Underline indicates the test set is resampled to match the training set. mIoU: higher is better.}
  \vspace{1ex}
    \begin{tabular}{p{4.5cm}|cc|cc|c}
%%%    \begin{tabular}{p{2.5cm}|cc|cc|c|c}
    \toprule
    % \multicolumn{7}{c}{Potsdam $\rightarrow$ Vaihingen} \\
    % \midrule
      Method   &  Training set &GSD & Test set &GSD & mIoU   \\
          \hline
    No domain adaptation &   \multirow{7}{*}{Potsdam}  &5cm & \multirow{7}{*}{Vaihingen}  &9cm  & 32.62  \\
    \underline{No domain adaptation}&  &5cm & &\textit{\underline{5cm}} &30.85  \\
    \underline{No domain adaptation}&  &\textit{\underline{9cm}} & &9cm &31.74  \\
  Standard domain adaptation &  &5cm & &9cm &  40.74\\
  \underline{Standard domain adaptation} & &5cm & &\textit{\underline{5cm}} & 41.77 \\
   \underline{Standard domain adaptation} & &\textit{\underline{9cm}}  & &9cm& 43.09 \\
  \textit{Our approach}  & &5cm & & 9cm  &\textbf{47.66} \\
    \bottomrule
    \end{tabular}
    \label{tab:resample}
    \vspace{-10pt}
\end{table*}
% \vspace{-5pt}
\subsection{Scale Attention Module}
\vspace{-5pt}
\label{subsec:sam}
The ASPP module in the DeepLab networks has demonstrated its effectiveness at handling multi-scale information for semantic segmentation. However, the input to the ASPP module is the low resolution feature maps which do not contain rich spatial information. We therefore adopt a self-attention mechanism to learn scale-enhanced feature maps for improved domain adaptation. 

We develop a scale attention module (SAM) to re-weight the multi-scale concatenated features as shown in Fig.\ \ref{fig:attention}. The input to our SAM consists of five feature maps extracted from different layers of an encoder network composed of a DeepLabV3+ model with a ResNet101 backbone. These feature maps are the outputs of each residual group and the final ASPP module. Adaptive pooling is used to transform the maps to the same spatial dimension. They are then concatenated into a single multi-scale feature map for performing self-attention.

The concatenated feature map ${f}\in \mathbb{R}^{H \times W \times C}$, where $H,W,C$ denote the height, width and number of channels, is first reshaped to $\mathbb{R}^{N \times C}$, where $N=H \times W$. The scale attention $\mathcal{A}({f})$ is then computed as
\begin{equation}
\setlength{\abovedisplayskip}{5pt}
\setlength{\belowdisplayskip}{5pt}
    \mathcal{A}({f})= \text{softmax} (\alpha({f})^{T} \beta ({f})).
\end{equation}
Here, $\alpha$ and $\beta$ are two $1 \times 1$ convolutional layers and $T$ indicates the transpose operation. The scale attention measures the impact of each channel based on various scales on all the other channels.

The final re-weighted feature map $\mathcal{O}({f})$ is computed using the scale-based attention weights through
\begin{equation}
\setlength{\abovedisplayskip}{5pt}
\setlength{\belowdisplayskip}{5pt}
    \mathcal{O}({f}) = \mathcal{A}({f}) \gamma({f}),
\end{equation}
where $\gamma$ is another $1 \times 1$ convolutional layer to transform the input. Finally, we reshape $\mathcal{O}({f})$ back to the original dimension $\mathbb{R}^{H \times W \times C}$ and feed it to the segmentation head for the final prediction. Note that we do not use the residual connection in our self-attention module since we only want the scale enhanced feature map.

We emphasize that our proposed SAM computes reweighted features along channels from different scales. The difference between the se-layer in Squeeze-and-Excitation Networks \cite{hu2018senet} and channel attention \cite{fu2019dual} is that the former uses only single-scale features while channel attention uses residuals to keep the original features. The goal of SAM is to enhance the features to include different scales in order to help the adversarial training when the scale discriminator is added.

% \section{Ablation studies}
% \label{sec:ablation}

\vspace{-5pt}
\section{Experiments}
\label{sec:experiments}
% \vspace{-3pt}
\subsection{Datasets}
\vspace{-5pt}
We evaluate our methods on three semantic segmentation datasets, two from the ISPRS 2D Semantic Labeling Challenge \cite{isprs_challenge} and a third from the DeepGlobe land cover classification challenge \cite{demir2018deepglobe}. 

\noindent \textbf{ISPRS 2D Semantic Labeling Challenge} This challenge includes two datasets, Vaihingen and Potsdam, both labeled with six classes: impervious surface, building, low vegetation, tree, car and clutter/background. The Vaihingen set contains 33 image tiles with size 2494 $\times$ 2064, of which 16 are fully annotated with class labels. The spatial resolution is 9 cm. We select five images for validation (IDs: 11, 15, 28, 30 and 34) and the remaining 11 for training, following \cite{maggiori2017high,sherrah2016fully}.
The Potsdam set contains 38 tiles with size 6000$\times$6000, of which 24 are annotated. The spatial resolution is 5cm. We select seven images for validation (IDs: 2\_11, 2\_12, 4\_10, 5\_11, 6\_7, 7\_8 and 7\_10) and the remaining 17 for training, again following \cite{maggiori2017high,sherrah2016fully}.

\noindent \textbf{DeepGlobe Land Cover Classification Challenge} This challenge introduces the first public dataset offering high-resolution sub-meter satellite imagery focusing on rural areas \cite{demir2018deepglobe}. It contains 1146 satellite images of size 2448$\times$2448, split into training/validation/test sets with 803/171/172 images. The images are from the DigitalGlobe Vivid+ dataset and have a pixel resolution of 50 cm. The classes include urban, agriculture, rangeland, forest, water, barren and unknown. The DeepGlobe dataset is more challenging due to its large coverage and dense annotations.
% \vspace{-10pt}

% \vspace{-4pt}
\subsection{Implementation Details}
\vspace{-5pt}
\label{subsec:implementation}

We implement our framework using the PyTorch toolbox \cite{paszke2017automatic} on a Titan V GPU with 12 GB memory. 

\noindent \textbf{Segmentation network} 
We choose the state-of-the-art semantic segmentation network DeepLabV3+ \cite{chen2018deeplabv3plus} with a ResNet101 backbone as our model architecture. The segmentation network $\mathcal{G}$ is trained using a Stochastic Gradient Descent (SGD) optimizer with Nesterov acceleration where the momentum is $0.9$ and the weight decay is $10^{-4}$. The initial learning rate is set to $2.5 \times 10^{-4}$ and is decayed using a polynomial decay policy with a power of $0.9$.

\noindent \textbf{Adversarial discriminator} 
We design our scale and feature discriminators using a fully convolutional network architecture following \cite{tsai2018learning}. The discriminator consists of 5 convolutional layers with $4 \times 4$ kernels and a stride of 2 as well as a padding of 1. Each convolutional layer is followed by a leaky\_ReLU activation unit with a negative slope of $0.2$. The channel number for each layer is set to 64, 128, 256, 512 and 1. The input to both discriminators is the predicted segmentation maps. To train the discriminators, we use the Adam optimizer \cite{kingma2014adam} with an initial learning rate of $10^{-4}$ and default momentum. We adopt the same polynomial decay policy as with training the segmentation network. We set the adversarial loss weights, $\lambda_f$ and $\lambda_s$, to 0.005.

\noindent \textbf{Evaluation metrics} 
We use segmentation metrics to evaluate land cover classification performance. In particular, we compute the per class Intersection over Union (IoU) and the mean over all classes (mIoU) as percentages (\%) where higher is better. We also compute the IoU gap with a model trained using the labeled target images where lower is better. This model serves as the oracle and can be considered an upper limit on the performance.

\begin{figure}[t]
    \setlength{\belowcaptionskip}{-10pt}
    \centering
    \includegraphics[width=\linewidth]{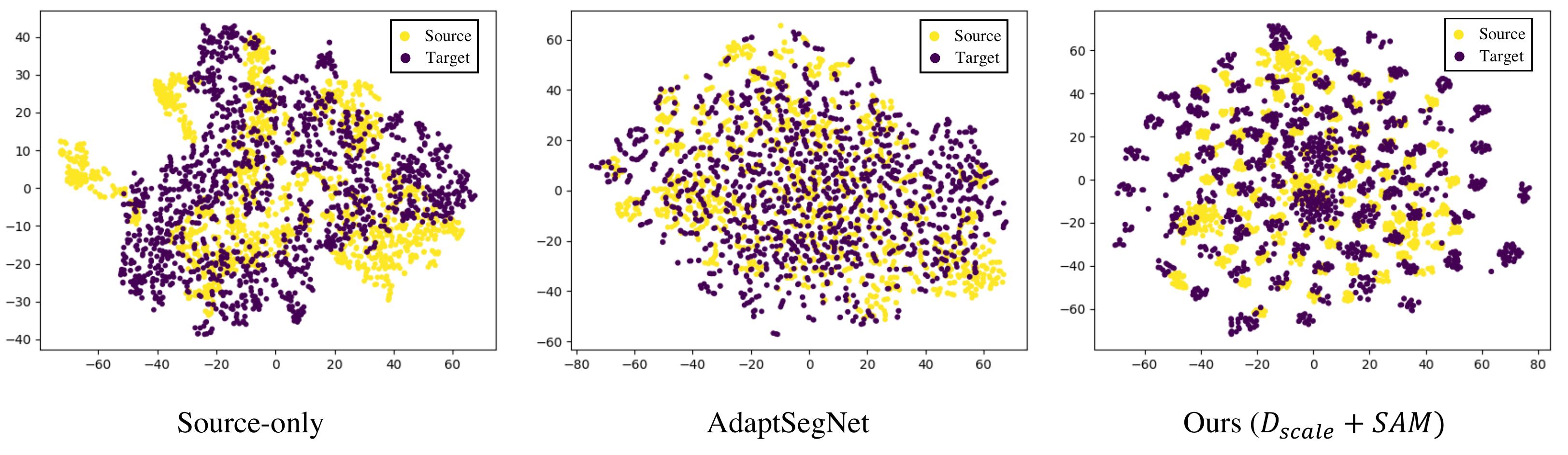}
    \caption{Visualization of the feature distributions using t-SNE. From left to right: before adaptation, domain adaptation NOT considering scale, domain adaptation considering scale. Source domain: ISPRS Potsdam, target domain: ISPRS Vaihingen. Our approach results in distributions that are more aligned and distinct. t-SNE hyperparameters are consistent between visualizations. (Zoom in to view details)}
    \vspace{-3pt}
    \label{fig:tsne}
\end{figure}

% Table generated by Excel2LaTeX from sheet 'Abaltion study'

% \vspace{-5pt}
\section{Experimental analysis}
\vspace{-5pt}
We evaluate the proposed framework as follows. We first compare our approach to standard methods for addressing scale mismatch such as image resampling and data augmentation (details in the supplementary material). This confirms our motivation. We then perform an ablation study to measure the contributions of the components of our framework. Finally, we compare our method to state-of-the-art to domain adaptation approaches.
%%%We evaluate the proposed framework by the following steps. We first compare our approach with some general baselines, data resampling and data augmentation. The goal is to verify that our motivation stands. Then An ablation study analysis follows. At last, we compare our method with multiple state-of-the-art methods. 
% \vspace{-5pt}
\subsection{General baselines}
\vspace{-5pt}
\noindent\textbf{Scale adaptation (ours) v.s. image resampling} A simple yet reasonable approach to address scale mismatch between source and target domains is simply to resize the resolution of one to match the other. We experimented with this using bilinear interpolation. (This assumes that the resolution of each is known which is our case.) It turns out, perhaps surprisingly, that this is not effective if there are other domain shifts such as location. Table~\ref{tab:resample} compares the results between resampling and our proposed method. The second and third rows show the results of training on 5cm Potsdam imagery and then testing on 9cm and 5cm (resampled) Vaihingen. The fourth through sixth rows incorporate standard non-scale aware domain adaptation, AdaptSegNet \cite{tsai2018learning}.  The results show some improvement, but they are still inferior to our method shown on row seven which achieves an mIoU of 47.66\% \emph{without resampling the target dataset}. This confirms that scale and other sources of domain mismatch are not necessarily independent and should be handled jointly. Similar results for Potsdam$\rightarrow$Vaihingen can be found in the supplementary materials.

\noindent\textbf{Impact of $D_{scale}$} We further analyze the proposed $D_{scale}$ by visualizing the feature representations using t-SNE \cite{maaten2008tsne}. As shown in Fig. \ref{fig:tsne} (left), given two datasets with different scales, the high-dimensional feature representations from the encoder are not aligned when there is no adaptation. The source-only model shows strong clustering of the source domain, but fails to have similar influence on the target domain. After applying domain adaptation i.e. AdaptSegNet (middle), the feature representations are projected into an overlapping space. This is due to the effectiveness of adversarial learning which forces the two distributions to be aligned. However, the class clusters are barely visible in the target domain and so the model has difficulty identifying the class boundaries. With our adaptation (right), the features are not only aligned but the class clusters are distinct. This further supports our claim that scale and other sources of domain mismatch are not necessarily independent and are best handled together.

% \subsection{Cross-Scale and Single-Location}

% \begin{table}[t]
% % \setlength{\abovecaptionskip}{-2pt}
%     \setlength{\belowcaptionskip}{-3pt}
%   \centering
%   \caption{Ablation study of adapting from Vaihingen scale-1 to Vaihingen scale-1/2. $\mathcal{D}_{feat}$ and $\mathcal{D}_{scale}$ indicate the feature and scale discriminators. SAM is the scale attention module. w/o DA means no domain adaptation.}
%     \begin{tabular}{c|ccc|cc}
%     \toprule
%     \multicolumn{11}{c}{Vaihingen scale-1 $\rightarrow$ scale-1/2} \\
%     \midrule
%     Method & $\mathcal{D}_{feat}$  & $\mathcal{D}_{scale}$   & SAM  & mIoU & IoU gap\\
%     % \hline
%     % scale-1 &       &       &       & 50.21 & 17.43 \\
%     \hline
%  \multirow{2}{*}{w/o DA} &       &       &       &50.21 &17.43 \\
%     &    &       &   \checkmark     &  \textbf{53.75} &13.89\\
% \hline
%     \multirow{4}{*}{w/ DA } &   \checkmark     &      &       &  52.37 &15.27\\
%      &   \checkmark  &   \checkmark &  &54.75 &12.90 \\
%     &   \checkmark     &       & \checkmark      & 56.02 &11.62\\
   
%     &  \checkmark      & \checkmark      & \checkmark      & \textbf{57.29} &10.35\\
%     \hline
%     scale-1/2 &       &       &       &67.54  & 0\\
%     \bottomrule
%     \end{tabular}%
%   \label{tab:ablation}%
% \end{table}%
% \vspace{-10pt}
% \subsection{Ablation study and analysis}
% \vspace{-5pt}

\begin{table}[t]
  \centering
    \setlength{\belowcaptionskip}{-5pt}
   \caption{Ablation study of adapting from Vaihingen scale-1 to Vaihingen scale-1/2. $\mathcal{D}_{feat}$ and $\mathcal{D}_{scale}$ indicate the feature and scale discriminators. SAM is the scale attention module. w/o DA means no domain adaptation.}
   \vspace{1ex}
    \begin{tabular}{c|ccc|c|c}
    \toprule
    \multicolumn{6}{c}{Vaihingen scale-1 $\rightarrow$ scale-1/2} \\
    \midrule
    Method & $\mathcal{D}_{feat}$  & $\mathcal{D}_{scale}$   & SAM    & mIoU & IoU gap\\
    % \hline
    % scale-1 &       &       &       & 50.21 & 17.43 \\
    \hline
 \multirow{2}{*}{w/o DA} &       &       &       & 50.21 &17.43 \\
    &    &       &   \checkmark     & \textbf{53.75} &13.89\\
\hline
    \multirow{4}{*}{w/ DA } &   \checkmark     &      &       & 52.37 &15.27\\
     &   \checkmark  &   \checkmark & & 54.75 &12.90 \\
    &   \checkmark     &       & \checkmark      & 56.02 &11.62\\
   
    &  \checkmark      & \checkmark      & \checkmark      & \textbf{57.29} &10.35\\
    \hline
    scale-1/2 &       &       &       & 67.54  & 0\\
    \bottomrule
    \end{tabular}%
  \label{tab:ablation}%
  \vspace{-15pt}
\end{table}%
\label{subsec:ablation}

\begin{table*}[t]
  \centering
  \setlength{\belowcaptionskip}{-5pt}
  \caption{Bi-directional domain adaptation: Potsdam $\leftrightarrow$ Vaihingen. $^{*}$ indicates our DeepLabV3+ implementation. mIoU: higher is better. IoU gap: lower is better. Target-only serves as the oracle representing an upper limit on the performance.}
  \vspace{1ex}
    \begin{tabular}{p{3.5cm}|cccccc|c}
    \toprule
    \multicolumn{8}{c}{Potsdam $\rightarrow$ Vaihingen} \\
    \midrule
          & Imp. Sur. & Build. & Low vege. & Tree  & Car   & mIoU & IoU gap \\
          \hline
    Source-only & 22.85 & 52.57 & 21.56 & 46.72 & 19.39 & 32.62 & 38.41 \\
    ADDA$^{*}$ &  42.93	&50.91	&27.02	&30.18	&10.09	&32.23	&38.80 \\
    CyCADA$^{*}$ &  49.39 &  55.29  & 28.03  &  32.04 & 10.49  & 35.05 & 35.98 \\
    {AdaptSegNet$^{*}$ } & 53.72	&56.08	&24.74	&39.68	&29.49	&40.74	&30.28\\
    Ours ($\mathcal{D}_{scale}$)  & 55.24 & 58.23 & 26.64 & \textbf{50.87} & 36.45 & 45.49 & 25.54 \\
    Ours ($\mathcal{D}_{scale}$ + SAM) & \textbf{55.22} & \textbf{64.46}  & \textbf{31.34} & 50.40 & \textbf{39.86} & \textbf{47.66} & \textbf{23.37} \\
    \hline
    Target-only  & 77.75 & 86.32 & 59.81 & 72.81 & 58.44 & 71.03 & 0.00 \\
    \hline
    \midrule
    \multicolumn{8}{c}{Vaihingen $\rightarrow$ Potsdam} \\
    \midrule
    Source-only & 31.06 & 37.35 & 44.13 & 16.76 & 31.29 & 32.12 & 45.0 \\
  ADDA$^{*}$ &  41.33	&44.21	&36.07	&29.81	&15.11	&33.31	& 43.81 \\
    CyCADA$^{*}$ &  39.02     &  42.35     &  35.09     &   27.89    &  10.25     &  30.92     & 46.20 \\
    AdaptSegNet$^{*}$ & 45.81 & 41.97 & 46.08 & 35.35 & 37.42 & 41.33 & 35.79 \\

    Ours ($\mathcal{D}_{scale}$) & 49.36 & \textbf{47.08} & 51.49 & 37.17 & 39.91 & 45.00 & 32.12 \\
    Ours ($\mathcal{D}_{scale}$ + SAM) & \textbf{49.76} & 46.82 & \textbf{52.93} & \textbf{40.23} & \textbf{42.97} & \textbf{46.54} & \textbf{30.58} \\
    \hline
    Target-only & 79.25&85.84&73.21&68.36&78.93&77.12&0.00
    \\
    \bottomrule
    
    \end{tabular}%
    \vspace{-15pt}
  \label{tab:pots_vai}%
\end{table*}%
To show the effectiveness of the proposed approach, we consider a simple scenario where two datasets are from the same location but with different scales. We 1) investigate how well standard domain adaptation can adapt for scale change even in isolation of location change, and 2) perform an ablation study of our framework. To do this, we use bilinear interpolation to resample images from ISPRS Vaihingen at 1/2-scale. The original images \textit{Vaihingen scale-1} have a GSD of 9cm and serve as the source dataset. The resampled images \textit{Vaihingen scale-1/2} have a GSD of 18cm and serve as the target. Table \ref{tab:ablation} shows that standard domain adaptation with only a feature discriminator improves by $2.16\%$ mIoU over a non-adaptation baseline ($50.21\% \rightarrow 52.37\%$). Oracle performance, where the model is trained on the resampled images \textit{Vaihingen scale-1/2}, is $67.54\%$. The $15.27\%$ IoU gap between standard domain adaptation and the oracle demonstrates the limited ability of standard domain adaptation to handle scale variation. Table \ref{tab:ablation} also shows the individual contributions of our proposed scale discriminator ($\mathcal{D}_{scale}$) and scale attention module (SAM). Adding the scale discriminator results in an improvement of $2.38\%$ ($52.37\% \rightarrow 54.75\%$). Adding the scale attention module results in an improvement of $3.65\%$ ($52.37\% \rightarrow 56.02\%$). We find that $\mathcal{D}_{scale}$ and SAM are complementary. Combining both results in the highest mIoU, $57.29\%$. We also observe that our proposed SAM, as a feature enhancement technique, provides benefits even without domain adaptation. Incorporating just SAM into the baseline results in an improvement of $3.54\%$ mIoU ($50.21\% \rightarrow 53.75\%$).

In summary, this set of experiments shows that scale is intricately linked to the features in CNNs and that scale-specific adaptation and enhancement is important and advantageous.

% \vspace{-5pt}
\subsection{Comparison Study: Small Domain Gap}
\vspace{-5pt}
We now perform cross-location domain adaptation between Potsdam and Vaihingen. We consider this as a small domain gap scenario since both locations are in Germany and the difference in scale is moderate.  We compare our approach to three recent state-of-the-art domain adaptation methods for semantic segmentation, ADDA \cite{tzeng2017adversarial}, CyCADA \cite{hoffman2018cycada} and AdaptSegNet \cite{tsai2018learning}. For fair comparison, we implement our own versions ($*$) with a DeepLabV3+ segmentation network. Table \ref{tab:pots_vai} contains the quantitative results and Fig. \ref{fig:results_vai} shows the qualitative results. More qualitative results can be found in the supplementary materials.
% \vspace{-12pt}

 %Table generated by Excel2LaTeX from sheet 'cross-dataset'
\begin{table*}[t]
\setlength{\belowcaptionskip}{-3pt}
  \centering
   \caption{Bi-directional domain adaptation: DeepGlobe $\leftrightarrow$ Vaihingen. $^{*}$ indicates our DeepLabV3+ implementation. mIoU: higher is better. IoU gap: lower is better. Target-only serves as the oracle representing an upper limit on the performance.}
   \vspace{1ex}
% Table generated by Excel2LaTeX from sheet 'cross-dataset'
    \begin{tabular}{l|cccc|cccc}
    \toprule
          & \multicolumn{4}{c|}{DeepGlobe $\rightarrow$ Vaihingen} & \multicolumn{4}{c}{Vaihingen $\rightarrow$ DeepGlobe} \\
    \hline
          & Urban & R. land & Forest & mIoU  & Urban & R. land & Forest & mIoU  \\
    \hline
    Source-only & 14.29 & 13.60 & 0.00  & 9.30  & 0.17  & 1.05  & 2.04  & 1.09 \\
ADDA* &21.77	&19.38	&7.28	&16.14 &25.04&	3.97	&20.99		&	16.67\\
    CyCADA* & 20.38 & 20.39 & 3.48  & 14.75  & 24.37 & 3.29  & 24.30 & 17.32 \\

    AdaptSegNet* & 22.28 & 17.71 & 5.81  & 15.27  & 26.82 & 3.98  & 31.04 & 20.61 \\
    Ours($D_{scale}$) & 26.98 & 19.16 & 11.29 & 19.14  & 28.34 & 4.74  & 39.49 & 24.19  \\
    Ours($D_{scale}$+SAM) & \textbf{27.35} & \textbf{23.43} & \textbf{12.94} & \textbf{21.24}  & \textbf{29.02} & \textbf{5.42}  & \textbf{41.72}& \textbf{25.39} \\
    \hline
    Target-only & 82.04 & 59.81 & 72.81 & 71.55  & 68.09 & 28.77 & 75.06 & 57.31 \\
    \bottomrule
    \end{tabular}%
   \label{tab:deep_val}
\end{table*}%
\begin{figure*}[t]
    \centering
    \setlength{\belowcaptionskip}{-8pt}
    \includegraphics[width=\linewidth]{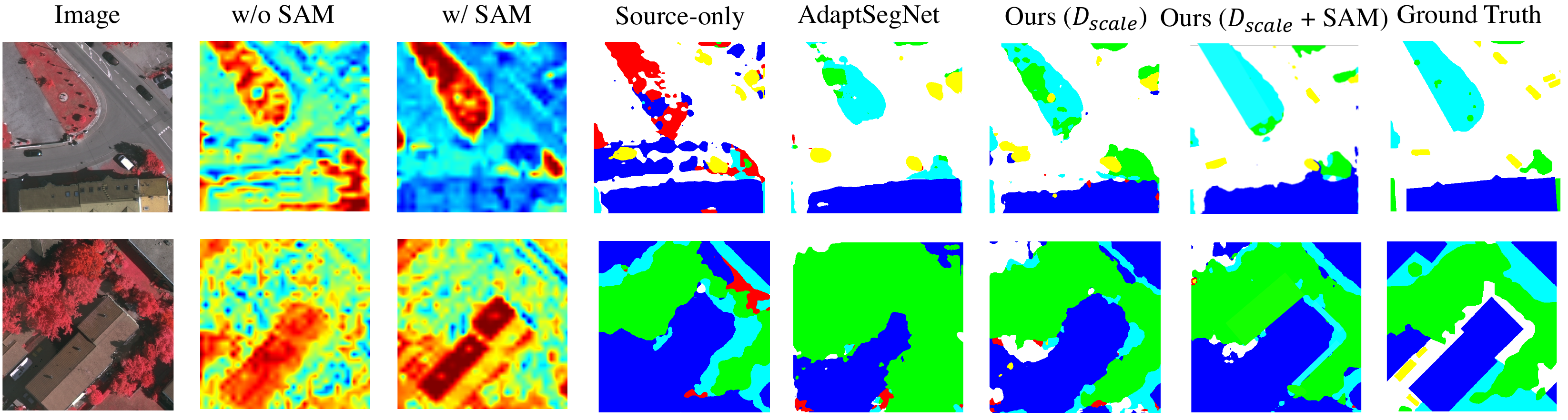}
    \caption{Visualization of Potsdam $\rightarrow$ Vaihingen. The second and third columns are the visualizations of channel feature maps, blue: low value, red: high value. Our class predictions have crisper boundaries and are more similar to the ground truth than standard methods. White: road, blue: building, cyan: low vegetation, green: trees, yellow: cars, red: clutter.}
    \vspace{-5pt}

\label{fig:results_vai}
\end{figure*}
We make several observations from Table \ref{tab:pots_vai}. First, standard domain adaptation does not work well on the cross-scale, cross-location problem. Take Vaihingen $\rightarrow$ Potsdam for example. CyCADA$^{*}$ performs even worse than the baseline, which is a model trained using the source dataset only without any domain adaptation. Second, our proposed scale adaptive framework achieves much higher performance in the cross-scale, cross-location scenario. With just our new scale discriminator, we improve over the previous best by $2.81\%$ ($42.19\%\rightarrow45.00\%$). Adding the scale attention module further boosts our accuracy to $46.54\%$, which is $4.45\%$ higher than AdaptSegNet$^{*}$ and $15.62\%$ higher than CyCADA$^{*}$. 

% \vspace{-2pt}
\noindent\textbf{Visualization of SAM} To further show the effectiveness of SAM, we visualize in Fig.\ \ref{fig:results_vai} the feature maps of the most weighted channel before (column 2) and after (column 3) applying SAM. We observe that SAM emphasizes feature maps at a more appropriate scale/size and results in better boundary estimates.

% \vspace{-2pt}
\noindent\textbf{Visual comparisons} Fig.\ \ref{fig:results_vai} shows visualizations of segmentation prediction maps adapting Potsdam to Vaihingen (columns 4-7). We see that the cross-scale, cross-location segmentation problem is very challenging for RS imagery. The results without adaptation are poor. Most predictions are clutter/background (incorrect). With standard domain adaptation, AdaptSegNet$^{*}$ generates improved but coarse predictions. Our framework results in more accurate predictions with crisper boundaries especially with SAM. Additional visualizations can be found in the supplementary materials.

% \vspace{-5pt}
\subsection{Comparison Study: Large Domain Gap}
\vspace{-5pt}
We now perform cross-location domain adaptation between DeepGlobe and Vaihingen. We consider this as a large domain gap scenario since the location varies significantly and the difference in scale is large (GSDs of 50cm and 9cm). Also, DeepGlobe is satellite imagery while Potsdam and Vaihingen are aerial imagery which introduces increased sensor and spectrum variation. To align the Vaihingen and DeepGlobe segmentation classes, we merge impervious surface and building as urban, match low vegetation to rangeland, and match trees to forest. The results of bi-directional domain adaptation between DeepGlobe and Vaihingen are shown in Table \ref{tab:deep_val}. We make several observations. First, due to the large domain gap, the source-only model performs quite poorly. For example, the DeepGlobe $\rightarrow$ Vaihingen mIoU is only $9.30\%$. Second, our scale adaptive framework again handles this now even more challenging scenario much better than standard domain adaptation. It achieves $5.97\%$ higher mIoU than AdaptSegNet$^{*}$ and $6.49\%$ higher mIoU than CyCADA$^{*}$. Similar improvements can be observed for Vaihingen $\rightarrow$ DeepGlobe.

% \vspace{-5pt}
\subsection{Discussion}
% \vspace{-5pt}
\vspace{-20pt}
\noindent \paragraph{Limitations} One limitation of our approach is that the scale of both the source and target domains must be known (or at least the scale difference). This is usually not a problem in RS image analysis since this information is typically available. We have also investigated methods to automatically estimate the spatial resolution of RS imagery using CNNs~\cite{8802954}. Another issue is that resampling images, using bilinear interpolation for example, is not equivalent to images whose scale varies due to having been acquired at different altitudes, etc. Despite this, our results show improved performance over standard domain adaptation. And, again, our primary goal is to illustrate the importance of scale in domain adaptation in RS segmentation. 
\vspace{-3pt}
\section{Conclusion}
\vspace{-5pt}
\label{sec:conclusion}
We establish that scale is a significant factor for domain adaptation in RS image segmentation. We propose a scale adaptive adversarial learning framework to address the challenging cross-scale, cross-location problem. This framework includes a new scale discriminator that explicitly adapts for differences in scale. It also includes a scale attention module to produce scale-enhanced features. Our framework outperforms standard domain adaptation methods, achieving state-of-the-art results for small (Potsdam $\leftrightarrow$ Vaihingen) and large (DeepGlobe $\leftrightarrow$ Vaihingen) domain gap scenarios.

\vspace{-3pt}
\section{Acknowledgement}
\vspace{-5pt}
This work was funded in part by a National Science Foundation grant, \#IIS-1747535. We gratefully acknowledge the support of NVIDIA Corporation through the donation of the GPU card used in this work.

{\small
\bibliographystyle{ieee_fullname}
\bibliography{egbib}
}

\end{document}